\documentclass[10pt,conference]{IEEEtran}
\ifCLASSINFOpdf
\else
\fi
\usepackage{amsmath}
\usepackage{gensymb}

\usepackage[pdftex]{graphicx}

\begin{document}

IEEE Copyright Notice
Copyright (c) 2017 IEEE
Personal use of this material is permitted. Permission from IEEE must be obtained for all other uses, in
any current or future media, including reprinting/republishing this material for advertising or promotional
purposes,  creating  new  collective  works,  for  resale  or  redistribution  to  servers  or  lists,  or  reuse  of  any
copyrighted component of this work in other works.\\

\noindent To be published in:\\

\noindent Proceedings of the 2017 International Conference on Computational Science
and Computational Intelligence (CSCI'17: 14-16 December 2017, Las Vegas,
Nevada, USA)\\
Publisher: IEEE Computer Society
Editors: H. R. Arabnia, L. Deligiannidis, F. G. Tinetti, Q-N. Tran, M. Qu Yang\\
ISBN-13: 978-1-5386-2652-8; BMS Part \# CFP1771X-USB\\
DOI 10.1109/CSCI.2017.313; paper acceptance rate: 21\%; 2017.\\
\newpage

%
\title{The Corpus Replication Task}

\author{\IEEEauthorblockN{Tobias Eichinger}
\IEEEauthorblockA{Service-centric Networking\\Telekom Innovation Laboratories\\
Technische Universit\"at Berlin\\
Berlin, Germany\\
Email: tobias.eichinger@tu-berlin.de}}


%


\maketitle

\begin{abstract}
In the field of Natural Language Processing (NLP), we revisit the well-known word embedding algorithm \emph{word2vec}. Word embeddings identify words by vectors such that the words' distributional similarity is captured. Unexpectedly, besides semantic similarity even relational similarity has been shown to be captured in word embeddings generated by \emph{word2vec}, whence two questions arise. Firstly, which kind of relations are representable in continuous space and secondly, how are relations built. In order to tackle these questions we propose a bottom-up point of view. We call generating input text for which \emph{word2vec} outputs target relations solving the Corpus Replication Task. Deeming generalizations of this approach to any set of relations possible, we expect solving of the Corpus Replication Task to provide partial answers to the questions.\\



\end{abstract}

%
\IEEEpeerreviewmaketitle

\section{Introduction}

Extracting meaning in the form of intra-word similarities from textual data and representing the latter in a continuous vector space has many useful applications in the field of NLP such as word completion, text classification, text generation, or sentiment analysis. However, there seems to be a natural limit to the \emph{degree of similarity} \cite{Ju12}, \cite{Mi13c} we are able to extract from natural human-generated text. We suggest that artificially generated text is equally suited for boiling down meaning to vectors. The paper at hand presents work in progress on the following questions. Firstly, which kind of relations are at all representable in continuous space and secondly, how are relations built in the word embedding algorithm \emph{word2vec}. Our intial findings support the view that the generation of text resulting in basic relations (such as [king:man] = [queen:woman]) may be mixed to sophisticated text generation schemes which produce any set of relations. We call this procedure \emph{Corpus Replication Task}.

\subsection{Extracting meaning from text}
The idea of discovering linguistic meaning in the structure of textual data dates back to the early 20$^{\mathrm{th}}$ century, when linguists like Leonard Bloomfield \cite{Bl65} or Ferdinand de Saussure \cite{Sau66} paved the way toward what would later be called \emph{structuralism}. The idea that structure itself was already key to the \emph{meaning} of linguistic entities has been picked up by Zellig Harris \cite{Ha54}, which nowadays may be condensed into the widely known \emph{Distributional Hypothesis}.
\\

\noindent\textbf{Distributional Hypothesis\footnote{For excellent reviews on the history and theoretical computational background of the distributional hypothesis, the gentle reader is recommended to confer to both M.Sahlgren's \cite{Sa06},~\cite{Sa08} and J.R.Curran's \cite{Cu03} works.}:}
\begin{quote}
	Words with similar distributional properties have similar meanings.\\
\end{quote}

A common ground for the analysis of distributional properties are word embeddings. Word embeddings identify words by vectors preferably such that the words' distributional similarity is captured. Yet the nature of distributional similarity is intrinsically vague and comprises many different aspects. They are commonly condensed into \emph{paradigmatic} and \emph{syntagmatic} similarity\footnote{Also known as \emph{rapport associatif} and \emph{rapport syntagmatique} in the original French version by de Saussure \cite{Sau68}.} in linguistics. Magnus Sahlgren reformulated the Distributional Hypothesis.\\

\noindent\textbf{Refined Distributional Hypothesis (\cite{Sa06},~\cite{Sa08})}:
\begin{quote}
	A word-space model\footnote{Discrete word embeddings with co-occurence word vectors are often referred to as word-space model in the field of linguistics.} accumulated from co-occurrence information contains syntagmatic relations between words, while a word-space model accumulated from information about shared neighbors contains paradigmatic relations between words.\\
\end{quote} 
Syntagmatic information is given in collocations such as 'hermetically sealed', where 'hermetically' very seldomly appears without 'sealed'. Paradigmatic information is given for example in word substitutes such as 'apple' and 'pear'. They are commonly used interchangeably, which is equivalent to the fact that they have equal contexts.

\subsection{Context in textual data}

Textual data are always \emph{sequential}. Every word in a text has a determined predecessing and succeeding word (unless they are either the first or last word of the text). Hence we may regard text as a (discrete) time series with realizations in the vocabulary space\footnote{When learning word embeddings we shall omit any possible meaning implied by punctuation.}. A Markovian point of view would suggest the past $n$ words $\{w_{t-n},..,w_{t-1}\}$ to yield knowledge of the successor $w_t$. Such subsequences are often referred to as \emph{left context}, analogously the future $n$ words $\{w_{t+1},..,w_{t+n}\}$ are referred to as \emph{right context}. In \emph{word2vec}, the union of left and right contexts are simply called context, where the integer $n$ is called the window size. Findings in Mikolov et al. \cite{Mi13a}, \cite{Mi13b} suggest weighing context words closer to $w_t$ higher than distant ones and choosing a window size of $n=5$ for better performance of \emph{word2vec}. The implementation of context is a controlling factor for the resulting word embeddings.

Subsampling the input text by dropping rare words, reducing frequent words or varying the window size provide tools of tuning word contexts. Apart from those methods, one may use \emph{dependency-based contexts} \cite{Le14} or \emph{growing contexts} as used in the Stochastic Memoizer \cite{Wo09}.

\subsection{Discovering similarity}
In \cite{Mi13b}, Mikolov et al. introduce the Continuous Bag-of-Words (CBOW) and Skip-Gram models. Both are Neural Network Language Models (NNLM). The former predicts a word, which fits a given context, whereas the latter predicts the context, which fits a given word.
\begin{figure}[htbp]\label{fig:1} 
	\centering
	\begin{eqnarray}
	CBOW :& \textrm{ \{fruit, an, is, a\}} \rightarrow \textrm{apple}\\
	Skip-Gram:& \textrm{apple}\rightarrow \textrm{\{fruit, an, is, a\}}
	\end{eqnarray}
	\caption[LOF-compModels]{Simplified comparison of the CBOW and Skip-Gram approaches.}
\end{figure}

Although both approaches seem very similar, they have been shown to perform quite differently on the \emph{Semantic-Syntactic Word Relationship Test} introduced by Mikolov et al. \cite{Mi13b}, wherein relations such as [country:capital] are considered \emph{semantic} and [adjective:adverb] are considered \emph{syntactic}. Since Skip-Gram simultaneously set new state-of-the-art benchmarks in both categories we will only consider \emph{word2vec} equipped with Skip-Gram in the sequel.

Note that syntagmatic similarity includes both semantic and syntactic similarity, since it constitutes co-occurence information. Paradigmatic similarity on the contrary comprises similarity with respect to shared contexts. An example of paradigmatically similar words are synonyms. They may be substituted at will and hence share the same contexts, whence they are paradigmatically similar. In continuous vector respresentations, both syntagmatic and paradigmatic similarity are usually measured with respect to the cosine distance, where a low cosine distance indicates high similarity. \emph{word2vec} finds paradigmatic similarity very well, yet struggles to surpass an accuracy level of about 65\% in syntagmatic similarity \cite{Mi13b}. 

\subsection{Increasing syntagmatic similarity}
One prominent approach of increasing syntagmatic accuracy in word embeddings are noise reduction schemes such as Noise Contrastive Estimation (NER) \cite{Gu12} or Negative Sampling \cite{Mi13a}. We would like to point out that the concept of noise in textual data is not explicitly defined in the literature. Yet, a majority of researchers agree that noise is present. We deem that this is  basically equivalent to the belief that a noise-free text will yield better syntagmatic accuracy levels. We rethink this idea by proposing a bottom-up approach. Assume that an artificially generated text is noise-free, then it will be able to produce a higher syntagmatic accuracy than human text. This motivates the introduction of the Corpus Replication Task.

\section{Reverse Engineering of Meaning}

\begin{figure}[!t]
	\includegraphics[width=3.8in]{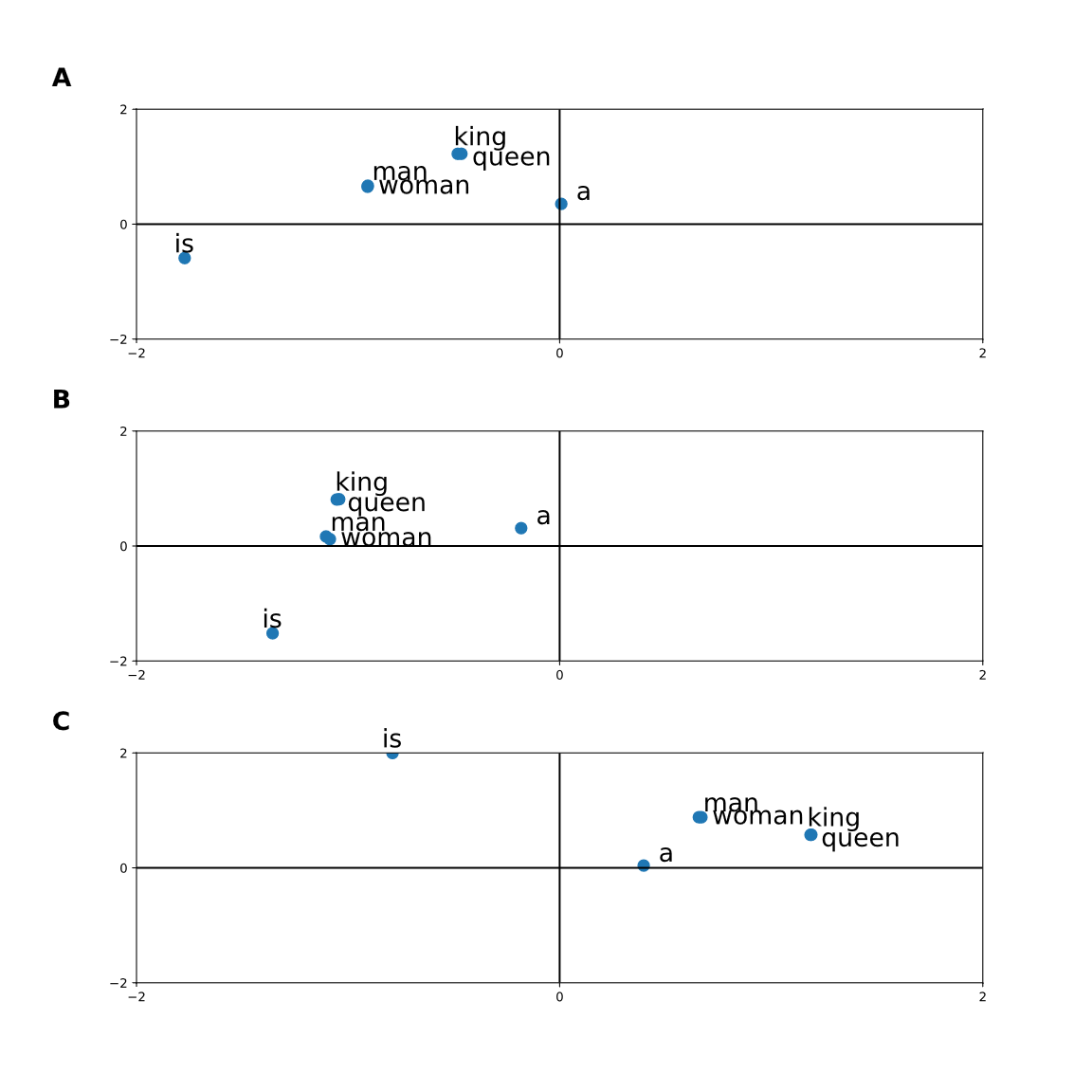}
	\caption{Learning the concept of royalty (cf. relation (\ref{eq:1})): 	\emph{word2vec} word vectors of \textbf{A}: 1.000, \textbf{B}: 10.000, \textbf{C}: 100.000 uniformly sampled and concatenated copies of the sentences \textbf{I} and \textbf{II} with window size $n=2$.}
	\label{fig_sim}
\end{figure}

In this section, we showcase the Corpus Replication Task. We shall give first results on reverse engineering a corpus of text for a given word relation. If \emph{word2vec} outputs a certain word relation $R$ for some input text $T$, we shall say that $T$ \emph{solves} $R$, or $T$ \emph{is a generative text for} $R$.

\subsection{Experiments}

We will solve the Corpus Replication Task for two basic syntagmatic word relations\footnote{ Note that we will measure syntagmatic similarity in terms of Euclidean distance instead of cosine distance as for example in \cite{Mi13a} or \cite{Mi13b}.}. We firstly specify a set of base sentences. Secondly, we define a probability distribution on the base sentences. Thirdly, we successively sample and concatenate base sentences into a text corpus. Finally, we run \emph{word2vec} on the generated text.
\subsubsection{Solving a syntactic word relation}
We shall learn the concept of \emph{'royalty'} from a randomly generated text, where for instance we define the concept of royalty abstractly as
\begin{eqnarray}\label{eq:1}
	vec(royalty) \approx& &vec(king)  - vec(man)\\
	\approx& &vec(queen) - vec(woman).\nonumber
\end{eqnarray}

Therefore, we define the following base sentences:
\begin{quote}
	\makebox[1.5cm]{\textbf{I:}} A king is a man.\par
	\makebox[1.5cm]{\textbf{II:}} A queen is a woman.\par
\end{quote}
We concatenate copies of \textbf{I} and \textbf{II} according to a Bernoulli distribution, i.e. $P($\textbf{I}$) = p$ and $P($\textbf{II}$) = 1-p$ for some $p \in (0,1)$. Then we run the \emph{word2vec} algorithm\footnote{We used the \emph{word2vec} implementation in the h2o framework version 3.14.0.1, cf. http://docs.h2o.ai/h2o/latest-stable/h2o-docs/data-science/word2vec.html for details.} with window size $n=2$ and calculate two-dimensional word vectors. The results for $p=0.5$ are shown in Fig. 2. We observe that the number of copies in the input texts does not change the word representations qualitatively. Furthermore, we see that the vector pairs ($vec(man)$,  $vec(woman)$) and ($vec(king)$, $vec(queen)$) clump. They are paradigmatically similar due to the fact that both are being used as substitutes of each other in the sentences \textbf{I} and \textbf{II}. As a consequence we have
\begin{eqnarray*}
vec(king) - vec(queen)& \approx 0,  \\
vec(man) - vec(woman)& \approx 0,
\end{eqnarray*}
which is equivalent to solving relation (\ref{eq:1}).

\subsubsection{Solving a semantic word relation}
A very commonly mentioned semantic word relation is the following:
\begin{equation}\label{eq:2}
vec(germany) + vec(capital) \approx vec(berlin).
\end{equation}
We shall build a generative text for (\ref{eq:2}) with the following sentences:
\begin{quote}
	\makebox[1.5cm]{\textbf{III:}} Berlin is the capital of Germany.\par
	\makebox[1.5cm]{\textbf{IV:}} Germany has a capital.\par
	\makebox[1.5cm]{\textbf{V:}} Berlin is the capital.\par

\end{quote}
We then uniformly sample sentences \textbf{III}, \textbf{IV}, and \textbf{V} and concatenate them into a text. The resulting word representations are presented in Fig. 4. For an input text size of 1.000 sentence copies, $vec(capital) + vec(germany)$ is very far from $vec(berlin)$, $vec(the)$ being the closest with respect to the Euclidean distance. However, increasing the input text size to 10.0000 copies shows that (\ref{eq:2}) is solved. For 100.000 copies, (\ref{eq:2}) is still solved, yet $vec(the)$ approaches $vec(berlin)$ again.

\begin{figure}[!t]
	\includegraphics[width=3.8in]{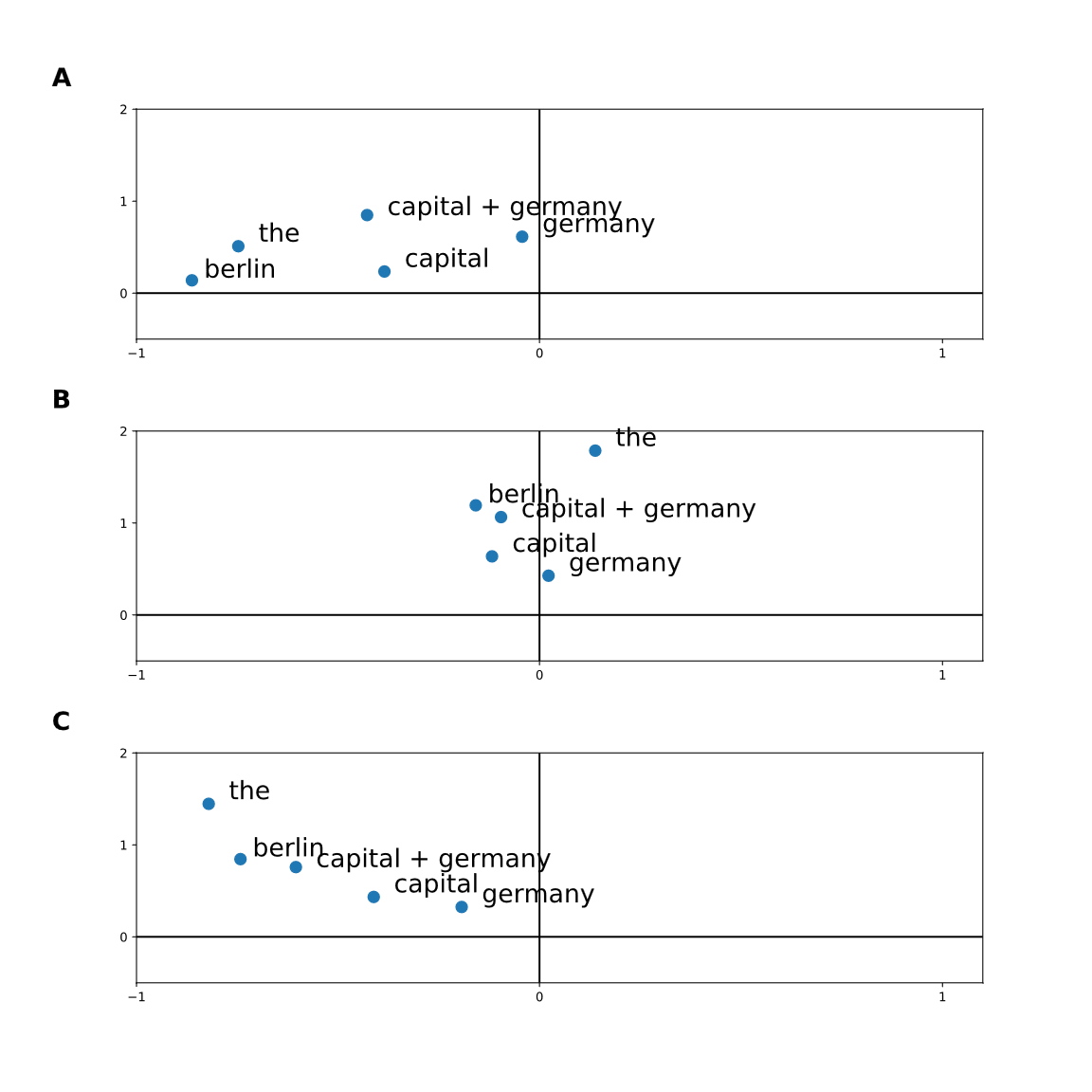}
	
	\caption{Learning the capital of Germany: 	\emph{word2vec} word vectors of \textbf{A}: 1.000, \textbf{B}: 10.000, \textbf{C}: 100.000 uniformly sampled and concatenated copies of the sentences \textbf{III}, \textbf{IV} and \textbf{V} with window size $n=2$. Word vectors for \emph{a}, \emph{has}, and \emph{is} sprayed into the periphery of the plot.}
	\label{fig_sim4}
\end{figure}

\subsection{Adjusting the window size}

In the above solutions, we used a window size of $n=2$. This choice is not arbitrary. It perfectly fits the distance in which the words we would like to relate to each other are positioned in the sample sentences. For example, if we want to learn a relation via \emph{word2vec}, we have to guarantee that the contexts of both words coincide enough. Take for instance a concatenation of 10.000 copies of the sentence \emph{'Berlin is the capital of Germany'} and set the window size $n=2$. Bundling words with equal contexts, we obtain the pairs $\{berlin, capital\}$, $\{is, of\}$, and $\{the, germany\}$. The \emph{word2vec} output shows that indeed vectors of the same set point into the same direction, whereas vectors of distinct sets span an angle of $120\degree$. In other words, vectors of the same set are (maximally) paradigmatically similar, whereas vectors of distinct sets are (maximally) paradigmatically dissimilar. We will call such vector sets \emph{contextually independent}.

Note that decreasing the window size to $n=1$ causes word vectors to spray due to less context overlap, or equivalently each vocabulary word now shares less neighbors with other vocabulary words. On the contrary, increasing the window size to $n=3$ causes all vectors to crumple up due to coinciding contexts. Hence, the concept of contextual independence directly depends on the choice of window size and the composition of base sentences.

\subsection{Towards meaning in higher dimensions}

Up until now we have only considered word embeddings in two-dimensional space. However, with increasing vocabulary size a generalization to higher dimensional space seems inevitable in order for word representations to capture larger sets of word relations. As pointed out in the previous subsection, there are sets of word vectors which are contextually independent from each other (for a fixed window size). In two-dimensional space, it is not possible to embed four vectors with pair-wise equal cosine-distance. However, this is feasible in three-dimensional space. We believe that it is possible to glue together low-dimensional word representations into higher-dimensional ones, which is part of our future research.

\subsection{Imbalanced corpora and complex distributions}

We also conducted the above experiments with non-uniformly sampled input texts. For example solving (\ref{eq:1}) was still feasible for very small probabilites $p = P($\textbf{I}$)$. The results are shown in Fig. 4. We observe that increasing $p$ causes the vector pairs $vec(man)$, $vec(woman)$ and $vec(king)$, $vec(queen)$ to collapse. We found very similar behavior for small values of $p_{1} = P($\textbf{III}$)$. Therefore, the extraction of word relations seems to be rather robust with regard to changes in the probability distribution. The evaluation of more complex distributions - in particular with strong conditional properties - is up for future research on the matter.

\begin{figure}[!t]
	\includegraphics[width=3.8in]{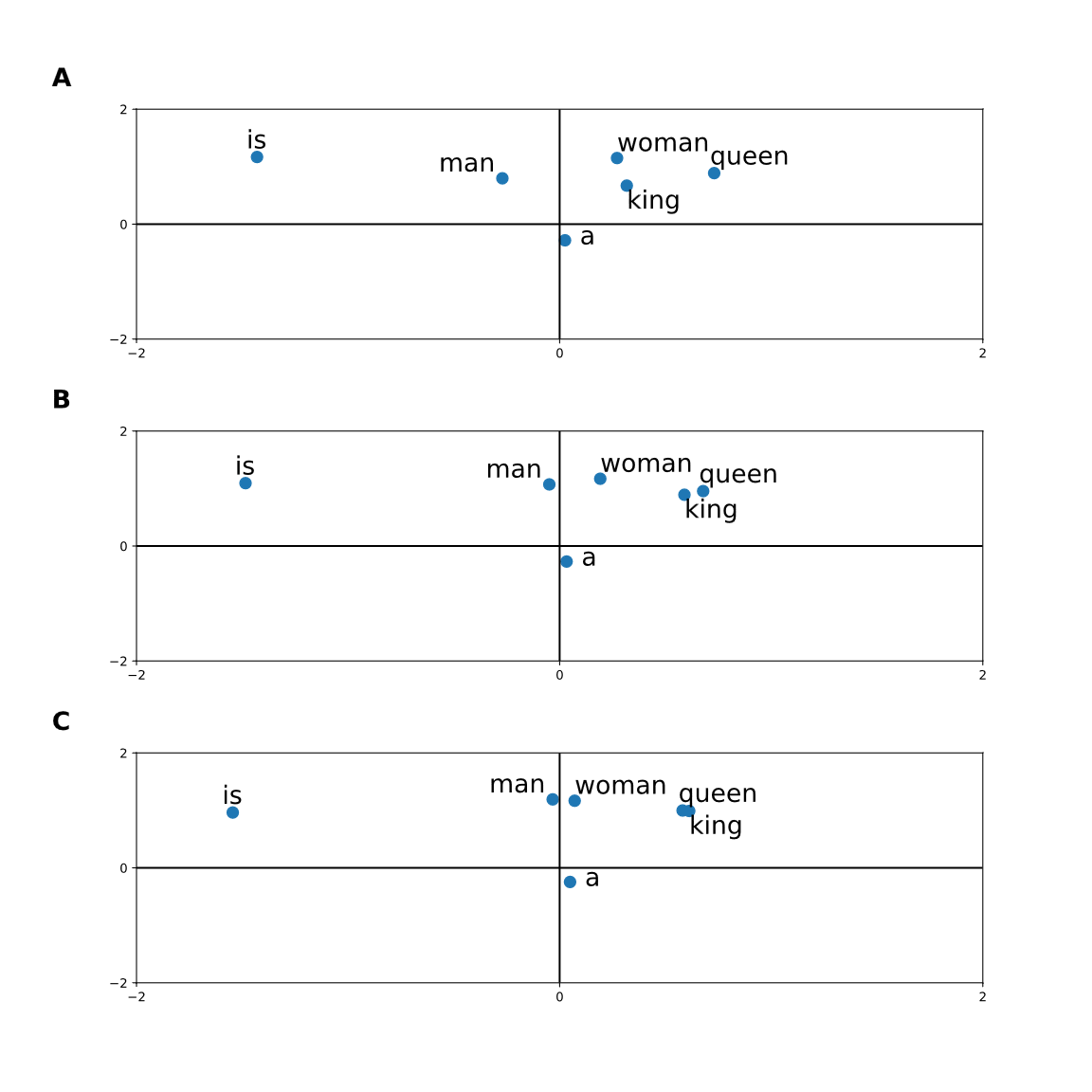}
	
	\caption{Learning relations non-uniformly: \emph{word2vec} word vectors of 10.000 binomially sampled and concatenated copies of the sentences \textbf{I} and \textbf{II} with window size $n=2$ and \textbf{A}:$P($\textbf{I}$) =0.002$, \textbf{B}:$P($\textbf{I}$)=0.005$, \textbf{C}:$P($\textbf{I}$)=0.01$.}
	\label{fig_sim2}
\end{figure}
\section{Related Work}

An excellent summary on the computational history of distributed word representations and linear regularity is given in \cite{Mi13c}. As this paper presents a reversed point of view on the matter, we propose further readings in the design of context. In \cite{Le14}, Levy and Goldberg present a dependency-based definition of context including a comparison to a non-Markovian definition of context as used in the Stochastic Memoizer \cite{Wo09} introduced by Wood et al. For a review on the extraction of relations of higher degree see for example \cite{Ju12}, \cite{Mi13c}. A comparison of the CBOW and Skip-Gram methods in \emph{word2vec} to other word embedding algorithms is provided in the papers \cite{Mi13a} and \cite{Mi13b} by Mikolov et al. \emph{word2vec}'s functioning is explained in great detail in \cite{Ro14} and \cite{Go14} .

\section{Conclusion}
We presented a view on distributional meaning of words from both a linguistic and computational point of view, alluding to the idea that artificially generated text is equally capable of representing in particular syntagmatic meaning. 
We introduced and solved the Corpus Replication Task for basic syntagmatic word relations such as $vec(germany) + vec(capital) \approx vec(berlin)$ and $ vec(king) - vec(man)\approx vec(queen) - vec(woman)$ in two-dimensional space. Solving the Corpus Replication Task in higher dimensions is most probably feasible by sticking together two-dimensional solutions. We believe that solving the higher dimensional case will shed light on the meaning extraction process of \emph{word2vec} by revealing opportunities and limitations in general. This, however, is up for future work.

\bibliographystyle{IEEEtran}
\bibliography{IEEEabrv,ISBD17} 

\end{document}